# A hybrid entity-centric approach to Persian pronoun resolution

Hassan Haji Mohammadi [a,1], Alireza Talebpour [b], Ahmad Mahmoudi Aznaveh [b], Samaneh Yazdani [a]

[a] Islamic Azad University Tehran North Branch, Tehran, Iran

[b] Shahid Beheshti University, Tehran, Iran

## Abstract

Pronoun resolution is a challenging subset of an essential field in natural language processing called coreference resolution. Coreference resolution is about finding all entities in the text that refers to the same real-world entity. This paper presents a hybrid model combining multiple rule-based sieves with a machine-learning sieve for pronouns. For this purpose, seven high-precision rule-based sieves are designed for the Persian language. Then, a random forest classifier links pronouns to the previous partial clusters. The presented method demonstrates exemplary performance using pipeline design and combining the advantages of machine learning and rule-based methods. This method has solved some challenges in end-to-end models. In this paper, the authors develop a Persian coreference corpus called Mehr in the form of 400 documents. This corpus fixes some weaknesses of the previous corpora in the Persian language. Finally, the efficiency of the presented system compared to the earlier model in Persian is reported by evaluating the proposed method on the Mehr and Uppsala test sets.

## 1. Introduction

Coreference resolution is an essential subtask of natural language processing. Its goal is to cluster all mentions in text that refer to the same real-world entity. Coreference resolution has many applications in other fields of natural language processing, such as information extraction, summarization, machine translation, question and answer, and emotion analysis [1-9]. Coreference resolution includes two separate parts of mention detection and mention clustering. In mention detection, all in-text mentions are extracted. In mention clustering, the extracted mentions are divided into coreference clusters.

Nowadays, the desire to research in the field of event-based coreference resolution is increasing among researchers. This field faces more challenges than coreference resolution. Recently, multiple deep neural network systems have been proposed in event-based coreference resolution [10-12]. Pronoun resolution is a challenging subset of coreference resolution. In pronoun resolution, a suitable antecedent is selected for each anaphoric pronoun in the text. Pronoun

---

**1** Email addresses : *h.hajimohammadi@iau-tnb.ac.ir* (Hassan Haji Mohammadi), *Talebpour@sbu.ac.ir* (Alireza Talebpour) , *a_mahmoudi@sbu.ac.ir* (Ahmad Mahmoudi Aznaveh), *samaneh.yazdani@gmail.com* (Samaneh Yazdani) resolution has several usages in other fields of natural language processing, especially in machine translation.

Most coreference and pronoun resolution systems have been developed in English and Chinese. Researchers in languages with less effort in this field (such as Persian) face more challenges due to the lack of pre-processing resources and suitable corpora. In this article, the authors use the Persian language as a research direction. Due to the structural difference between Persian and languages like English, a model that is accurate in this language may not have a good result in English and vice versa.

Coreference resolution systems are divided into rule-based and machine-learning systems [13]. Early rule-based coreference systems have used a series of hand-written rules for mention clustering [14-19]. The sieve-based system was one of the most successful rule-based coreference systems [20-22]. This system consists of a series of rule-based sieves, which are ordered from the most accurate to the least accurate. The architecture of this system is entity-based and includes the extraction of rich features.

One of the advantages of rule-based systems is their simplicity. Another strength of these systems is that they can extract entity-level features based on the entire document [23]. However, rule-based systems lonely can not solve the problem of coreference resolution. One of the most important disadvantages of rule-based systems is that they are unsuitable for pronoun resolution. Pronoun resolution requires strong inference by combining different features, which machine learning models conduct this feature conjunction. The results of this research prove the truth of this claim.

Machine learning systems focus on building models automatically from training data. One of the properties of machine learning systems is the ability to work with rich semantic features. Therefore, these models are suitable for developing pronoun resolution systems [24]. In a category, machine learning systems are divided into four classes. Mention-pair, mention-ranking, entity-based and entity-mention models. The main issue with the mention-pair and mention-ranking models is that the mentions merge decision is taken locally. To solve this issue, researchers have developed entity-based machine-learning models. Entity-based models also have some problems. One of the most important problems of entity-based models is that their computational complexity is high for large documents.

Recently, end-to-end models have had a good performance in coreference resolution systems. These models do not need a processing pipeline and can perform all the steps of coreference resolution jointly. Although these algorithms add appropriate context knowledge from an external unlabeled data source to the model, this information is insufficient to cover all the knowledge needed to construct a pronoun resolution system [25].

Other challenges of end-to-end models include the following: large documents with large coreference chains, the need for a large number of training examples, and the problem of gender bias [23]. Despite the recent advances, pronoun resolution has still been challenging. One of the most important reasons is that external knowledge cannot be easily extracted through training samples [26]. Effective integration of contextual information and external knowledge is one of the challenges in the field of pronoun resolution.

This article performs the expriements on two corpora, RCDAT [27] and Mehr corpus. The authors develop the Mehr corpus in this paper. The details of the Mehr corpus are given in section 3. All the challenges in end-to-end models can be handled using a proper combination of rule-based and machine-learning models, so this article uses a hybrid model. The architecture used in the presented system is a hybrid architecture. This architecture consists of several rule-based sieves and one machine-learning sieve for pronoun resolution. The machine learning sieve uses a combination of hand-extracted features and word vectors. The presented model is an entity-centric model . Entity-centric means that the architecture is based on the entity model and builds partial entities incrementally. In this architecture, adding cluster-level features has increased the efficiency of the pronoun resolution system.

The introduced model using random forest (and above MLP neural network) has had good results. One of the reasons is that the random forest can model a large number of conjoined features. The authors implemented all sieves as processing pipelines to make the model as simple as possible. In the system's output, a suitable antecedent is found for each anaphoric pronoun.

The remainder of this article is structured as follows:

Section 2 analyzes related works to coreference resolution in Farsi and English. In Section 3, the existing coreference resolution corpora in Persian are reviewed. Also, in section 3, the details of the Mehr corpus are described. Section 4 covers the hybrid model presented in this article. Section 5 reports all experiments and results performed on RCDAT and Mehr Corpus. Finally, the paper's conclusion is presented in Section 6.

## 2. Related work

In this section, first, related works to coreference resolution systems in English are reviewed. Then, coreference and pronoun resolution systems in Persian will be examined.

**2-1 Coreference resolution systems in English**

The system presented in [22] can be mentioned among the successful rule-based systems in coreference resolution. This system performs coreference using a series of rule-based sieves arranged in order of decreasing precision. In this system, early sieves have more precision. The next sieves are less accurate but increase the system's recall. This system has been used in Stanford University preprocessing tools [28], and to date, various languages have used this framework to coreference resolution systems. For example, in the article proposed in [29], this framework is used in Indonesian.

The early coreference machine-learning systems were mention-pair models. In this model, a binary classifier for each mention and antecedent candidate decides whether they are coreference or not. Contrary to the simplicity of this model, many coreference systems have used this architecture in their architecture [30-34]. One of the most critical weaknesses of this model is that mentions are considered locally. The mention-ranking model, for each mention, considers a suitable candidate among several candidates [35, 36]. In this model, an antecedent may not be selected for a mention, which means that the mention is non-anaphoric. Many current deep-learning models have used this architecture in their systems [37, 38].

As mentioned in the previous paragraph, one of the most significant problems of the mention-pair models is that they examine features locally. Researchers have proposed the entity-based model [39-41] to solve this problem. In this model, each mention (or partial chain) is compared with a previous chain (or partial chain) to determine if it is a coreference relation or not. This model defines the features at the cluster level. The cluster-ranking model is created from the combination of the entity-based models and the ranking models. In this model, the ranking phase is performed at the cluster level. That is, previous partial chains are scored, and the highest score is considered as chain antecedent.

Hybrid systems have taken advantage of both rule-based and machine-learning models. The authors of the article [24] presented a hybrid system of machine learning and rule-based sieves. This system uses one machine learning sieve for each entity type. Compared to the sieve model presented in [22], the advantage of this model is that the number of sieves is reduced, and a distinct sieve is used for each entity type. This article also investigates the effects of using a dependency tree instead of a structural parser. In addition, attempts have been made to develop a hybrid system in other languages. For example, we can refer to the hybrid system presented in [23].

Recently, End-to-end systems have increased coreference systems' efficiency by eliminating pre-processing steps and manual feature extraction. These models perform both mention detection and mention clustering jointly. The article's authors [38] produced the embedding vector of the spans within the sentences by averaging a bidirectional LSTM network. The architecture of this model is mention-ranking. This model considers spans locally. The authors of the following articles have increased the efficiency of the end-to-end model by adding the global term to this model. The proposed paper [42] improved the span-ranking architecture using the biaffine attention mechanism. By using the mention-ranking and reference-ranking in the loss function, the authors optimized both simultaneously. Lee et al. in paper [43] increased the efficiency of the end-to-end model by adding a higher-order concept to the model. Also, in this article, the computational complexity of the model has been reduced by using the antecedent pruning mechanism. Lai et al. have increased the efficiency of the baseline model [38] by introducing pre-trained transformer language in their paper [44] . Chai and Strube [45] introduced a new research direction by combining neural network models and discourse theory by adding the concept of centering theory to the neural network model.

Other articles implemented the end-to-end model's global term in another way. Kantor and Globerson [46] presented the entity equalization mechanism. In this method, each mention within a coreference chain is considered as an estimation of total mentions in that chain. Joshi et al. [47] Using contextual text embedding vectors such as BERT has increased the efficiency of end-to-end models. Using BERT instead of previous embedding methods such as ELMO, the authors have reported a suitable increase in efficiency. Next, Joshi et al. [48] introduced spanBERT by changing the Bert loss function. This system has significantly increased the efficiency of coreference systems. Miculicich and Henderson [49] proposed a graph model called GT2G for encoding coreference links in documents. In this model, all complete coreference graphs are created at once. The highest performance among coreference systems on the CoNLL 2012 corpus [50] is the BERT-based system presented by Wang et al. [51]. In this system, an actor-critic method is presented, which performs mention-detection and mention clustering jointly using reinforcement learning.

**2-2 Coreference resolution systems in Persian**

In this section, coreference and pronoun resolution systems in the Persian language are described. Moosavi and Ghassem-Sani [52] presented Persian's first pronoun resolution system. This system uses the ranking method to find the pronoun's antecedent in the PCAC-2008 corpus. The authors developed the PCAC-2008 corpus. This system founds the pronoun's antecedent in the corpus using the method and the features presented in the article [53]. The features used include pronoun features, antecedent features, and relational features. The authors compared their features with four models: max entropy, max classifier, perceptron, and c4.5. The results section reports the c4.5 classifier with a value of 44.7 F1 as the highest accuracy.

Hajimohammadi et al. [54] presented a new pronoun resolution corpus in Persian. The authors reported a 20% increase in efficiency by adding a new feature set to the paper [52] in their new corpus. As another effort, Nourbaksh and Bahrani [55] improved the results by applying a series of heuristics to solve the problem of an unbalanced dataset generated in the system [52] and adding semantic information. They have examined maximum entropy, perceptron, SVM, random forest, KNN, and C4.5 classifiers. The best-reported results were the decision tree, which was about 75 F1 on the PCAC-2008 corpus.

Fallahi and Shamsfard [56] presented a rule-based system to find the reference of pronouns. The system introduced in this article includes two parts: pre-processing and antecedent selection. In this article, using five simple rules, the antecedent of pronouns is determined up to three sentences before. The corpus used in this research is 5 Persian blogs, each of which has 20 pages. They compared their system with the system [52] and reported better accuracy and recall.

Nazaridoust et al. [57] presented Persian's first coreference resolution system. The authors developed the Lotus corpus and used 17 features to construct their coreference system. This system has been tested on 20% of the Lotus corpus using a decision tree, SVM, and NN classifiers. The best-reported results were related to neural networks with an accuracy of 39.4 F1 metrics.

The previous systems presented in Persian were all statistical works and on limited corpora. None of the previous systems have formed output chains, and they have not used main evaluation metrics such as conll to evaluate their system. With the development of RCDAT's one million tokens corpus [27] in Persian, the research direction in this language changed. The corpus developers reported 60 conll points by developing the mention-pair model using the SVM algorithm on the RCDAT corpus.

The following system that evaluated its results on this corpus was the system presented in [30]. This article uses a fully connected neural network to train the coreference classifier. A graph-based algorithm is used in constructing coreference chains, which is improved by 2.2 conll points compared to the previous system. In their following paper [31], the authors improved the results by maintaining the general framework of the model and only by changing the feature vector and adding a semantic feature to their article. The feature set used in this article includes three parts of heuristic features, word vectors, and semantic features introduced in [58]. In the output, coreference chains are created using the hierarchical clustering algorithm. Their reported result is 64.54 conll points on RCDAT.

## 3. Coreference corpora in Persian

In this section, coreference corpora in Persian are examined. This article uses two corpora to conduct experiments and report the results. The first corpus is the RCDAT corpus, which contains 1599 documents. The second corpus is the Mehr corpus developed by the paper's authors. The Mehr corpus contains 400 documents. In this section, the corpora in Persian have been examined. Then the details of Mehr's corpus development are given.

**3-1 Early corpora in Persian**

The first pronoun resolution corpus developed in Persian was PCAC-2008 [52]. This corpus contains 31 texts taken from BijanKhan's corpus [59]. This corpus annotated the closest antecedent for 2006 pronouns in the text. The Lotus corpus [57] is the first coreference resolution in Persian. This corpus contains 50 long texts from BijanKhan's corpus, labeled with coreference noun phrases.

**3-2 PerCoref Corpus**

This corpus [60] contains 212,646 tokens from 547 documents. These documents are randomly extracted from the web. This corpus uses three reference types direct, indirect, and no reference. Table 1 shows the statistics of this corpus.

**Table 1.** PerCoref statistics [60]

| | |
|---|---|
| **Number of documents** | 547 |
| **Number of tokens** | 212646 |
| **Number of sentences** | 6688 |
| **Number of paragraphs** | 4449 |
| **Average paragraph per document** | 8.13 |
| **Average sentence per document** | 1.5 |
| **Average token per document** | 31.79 |

This corpus labels six types of coreference relations (Identity, person/number suffix on the verb, Inferred, Event, Quantifier, and Cross-speech). Their abundance statistics are given in table 2.

**Table 2.** PerCoref corpus coreference relations

| Coreference relations | Count | Percent |
|---|---|---|
| Identity | 11812 | 58.4 |
| person/number suffix on verb | 3021 | 14.9 |
| Inferred | 2005 | 9.9 |
| Event | 1509 | 7.4 |
| Quantifier | 968 | 4.7 |
| Cross-speech | 901 | 4.4 |

### 3-3 RCDAT Corpus

This corpus [27] contains one million words and is published in the CoNLL format. The annotators label gold coreference, named entity, and animacy tags in this corpus. The corpus text includes various political, sports, and cultural topics. These texts are extracted from news sites. Table 3 describes the statistical information of the RCDAT corpus.

Table 3. RCDAT's statistical information

| | count |
|---|---|
| Number of documents | ۱۵۹۹ |
| Number of tokens | ۱٫۰۵۲۶۳۷ |
| Number of labeled tokens | ۴۰٫۸۱۳۸ |
| Number of entities | ۸۶۹۶۰ |
| Number of pronouns | ۲۶۸۷ |
| Number of proper nouns | ۳۴۱۲ |
| Number of animate entities | ۴۲۶۶ |

Each line of the corpus contains the following information: Name of the document, sentence number, token, pos tag (16 tags), named entity (13 golden tags), the root of the token or itself, original token, named entity (3 golden tags), index of the token in its coreference chain (gold ), coreference chain number (golden), animacy, phrase type, pos tag(100 tags).

### 3-4 Mehr corpus

This section describes the corpus developed by the authors. This corpus contains 400 texts from the Mehr news agency[2]. These texts are randomly extracted in different intervals. Two annotators tag each corpus document. This corpus is labeled by the preprocessing tool WebAnno[61]. This tool automatically generates the conll file for the labeled chains in its output. Figure1 shows the annotating steps by WebAnno. The corpus documents are divided into training, development, and testing. Table 4 shows the statistical characteristics of the Mehr corpus. The number of tokens in test documents is the same as the number of tokens in the test section of the MUC7 corpus [62]. About 13 thousand tokens have been selected.

Table 4. Statistical characteristics of Mehr's corpus

| Document | Document counts | Sentence counts | Token counts |
|---|---|---|---|
| Train and Development | 357 | 3189 | 158047 |
| Test | 43 | 286 | 13214 |

---

[2] Www.mehrnews.ir

| | | | |
|---|---|---|---|
| **Total** | 400 | 3475 | 171261 |

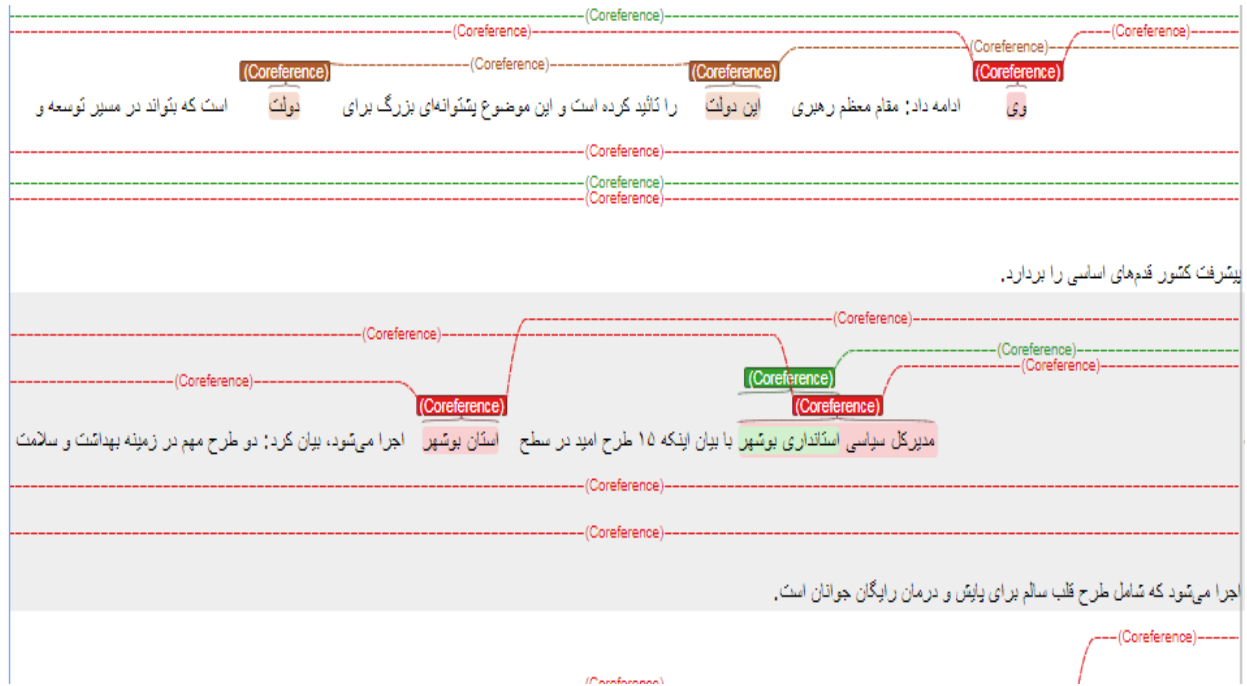

**Figure1.** A frame of WebAnno annotation tool

Mehr corpus has two types of labels.

- **Gold labels**

In this corpus, coreference labels for pronouns, noun phrases, named entities, and nested named entities are manually labeled. Only those participating in at least one coreference chain are labeled as mentions. Among the pronouns, some non-referential pronouns are also labeled, which are described below.

In Persian, some pronouns do not have a direct reference in the text. In fact, in Persian, some pronouns do not refer to anything. Non-referential pronouns are also labeled in the Mehr corpus.

Some pronouns, such as nobody and others, which do not have any reference in the text, are not labeled.

In Persian, some pronouns have an indirect reference. Their reference is the speaker of the quote, which may have been omitted in the document. In this corpus, these types of pronouns are also labeled. In the Mehr corpus, only pronominal singletons are labeled with coreference relations. The coreference relation type in this corpus is identity relation. We mention an example for each coreference relation in the Mehr corpus in the following.

  o **Coreference relations**
    - (محمد) به مدرسه آمد. بچه ها از دیدن (او) تعجب کردند.
    - *(Muhammad) came to school. The children were surprised to see (him).*

- (ابن سینا) دانشمند ایرانی است. (این دانشمند برجسته) در همدان مدفون است.
  - *IbnSina is an Iranian scientist. (This outstanding scientist) is buried in Hamadan.*

o **Indirect reference**
- (ما) باید به تیم ملی خود افتخار کنیم.
- (We) should be proud of our national team.

o **Non-referential pronouns**
- (این)، مساله مهمی است.
- (It) is an important issue.

- **System labels**

Text processing tools in Persian produce these labels automatically. These labels' accuracy differs from manual labels and depends on the accuracy of preprocessing tools. The automatic labels in the Mehr corpus include named entities, noun phrases, pos tags, syntactic groups, etc. This corpus uses the Hazm preprocessing tool[3], the comprehensive platform of intelligent content analysis[4], and the FarsiYar text preprocessing tool5. Table 5 shows the accuracy of preprocessing tools used in this corpus.

**Table 5**.Accuracy of preprocessing tools used in this paper

| Persian tools | Accuracy |
|---|---|
| Pos tagger | 97.1 |
| Shallow parser (Chunker) | 89.9 |
| Constituency parser | 85.2 |
| Named-entity tagger | 89 |
| Animacy recognition | 91 |
| Pronoun recognition | 96 |

## 4. Proposed method

The system presented in [31] has the highest efficiency among the systems presented in Persian. The authors compared their system by implementing the state-of-the-art English coreference resolution systems in the Persian RCDAT corpus. According to this comparison, their proposed system has higher efficiency. This paper improves the most efficient Persian system by combining rule-based and machine-learning models.

In this section, we will first introduce two basic pronoun resolution models. The hybrid system is proposed by combining these two models and defining the cluster-level features. The first baseline system is a sieve-based system, according to [22]. This baseline system develops rule-based sieves for Persian. The second baseline model is a mention-pair machine learning model. This model introduces hand-crafted features for pronouns, antecedents, and relational features. Finally, combining two baseline models and adding cluster-level and word embedding features introduces a hybrid entity-centric model. This paper shows that this hybrid model has the highest efficiency in Persian systems.

---

[3] https://www.sobhe.ir/hazm/
[4] http://catotalservices.com/UIHome/Index.aspx
[5] https://text-mining.ir/

**4-1 Persian sieve-based baseline model**

This architecture was introduced in [22] . The proposed architecture has been used in many peer-reviewed articles and off-the-shelves systems. This architecture consists of several sieves arranged in order of precision. Early sieves have more precision than following sieves. In each sieve, the mention is either merged into one of the partial clusters formed so far or finding an antecedent is left to the following sieves. In the current sieve and for the mention, the system does not consider only the local information of this mention with its previous mentions. This model considers the global information of previously partial clusters for merging with the current mention .

The architecture of this system is modular; that is, any new module can be easily added to the system.

In the introduced baseline system, eight sieves are used in Persian. Persian and English have many linguistic differences. Therefore, the sieves introduced in this article have differences from English sieves. Some English sieves have been removed, and some have been modified. Before applying sieves, documents are processed by existing NLP preprocessing tools. The system identifies the mention's boundary, which includes noun phrases, named entities, and pronouns. Also, mention-related features, such as number and animacy, are extracted. These mentions will be given as input to subsequent sieves.

- **Model architecture**

The architecture of the mention selecting is the same as the method introduced in the article [21]. The difference is that in Persian, unlike English, the document traversing direction is from right to left. The first issue to consider is the order in which mentions are selected for examination in each sieve. Because partial chains are incrementally formed in each sieve, selecting the representative mention of the chain is essential. Because in each sieve, only one mention participates as a representative of chains. The algorithm for selecting mentions in the current sieve is given as follows.

  o *Selecting mentions within the current sieve*

The partial chains created by the previous sieves are given as input to the current sieve. Only the first mention within the created partial chains is selected in the current sieve. For example, suppose for the current document, the partial chains $\{m_1^1, m_2^2, m_3^2, m_4^3, m_5^1, m_6^2\}$ has been constructed so far. In this example, the lower index is the mention number, and the upper index is the cluster number. In the current sieve, the model only tries to find the antecedent for $m_2^2$ and $m_4^3$ ($m_1^1$ is not checked because it is at the beginning of the document). The reason is that the first mentions of partial chains usually define the entity better, and the probability that they are pronouns or have prefixes and suffixes is less {Lee, 2013 #52}. Also, the first mentions of partial chains are usually closer to the beginning of the text and therefore have fewer candidates to consider as antecedents.

- *Antecedent selection for a current mention*

Each mention in the current sieve may be merged with a previous partial chain, or its antecedent may be assigned in the next sieve. For each mention in the current sieve, first, the mentions in the current sentence are sorted from right to left and checked in order. The document sentences are traversed from left to right if there is no answer in the current sentence for the current mention. Then in each sentence, the mentions are sorted from right to left to reach the answer. Sorting the list of candidates is important because the algorithm stops at the first confluence. Weak sorting causes the accuracy of the system to decrease. A sentence threshold window should be set in each sieve so the algorithm does not check more than that threshold limit.

- *Sharing features in the entity-centric model*

For each partial entity (which includes one or more mentions), the union operation is used on the mentions features (such as number, gender, and animacy). Then the result is shared for all mentions of that entity. If feature values differ for mentions of the entity, all its values are preserved. For example, the number detection module selects singular for the mention "group of students," and for the mention "five students," selects plural. If a partial entity is formed by merging two mentions of the previous sentence, it will have two singular and plural values for the number feature. Therefore, singular and plural pronouns can be added to this partial entity in the pronoun resolution sieve.

- **Model sieves**

This section describes the developed sieves for the proposed system.

- *Sieve 1 - Speaker sieve*

In this sieve, the pronouns in the quotation connect to their antecedent, which is the speaker of the quotation. The speaker of the quote must be identified first. The document specifies the speaker of the quotation in conversational texts. In non-conversational texts (like Mehr and RCDAT corpora), some simple rules are used to identify quoted verbs (such as said).

- *Sieve 2 – Exact string match*

In this step, if two strings match completely (this match includes all prefixes and suffixes), they will form a chain. This sieve has high precision.

- *Sieve 3 – Strict head match*

In head matching, stricter rules are used for merging two mentions. The reason is that two mentions may have the same head but not coreference. For example, the two mentions, "Tehran University "and "Washington University," have the same head but are related to two different named entities. In this step, if two mentions with the same head satisfy some strict conditions, they can be merged. These rules are given below.

1. If the head mention matches any head mention of the previous entities, we can check the following condition.
2. The mention's string must be an entire substring of its antecedent string. For example, the mention "Tehran Court" is a substring of the mention "Tehran High Court." Therefore, these two expressions can be merged under the satisfaction of these conditions.

- *Sieve 4 – Proper name match*

In this sieve, two mentions with the same proper noun head are merged in a partial chain if they satisfy some conditions. For example, the two mentions of "Beckham" and "David Beckham" merge and form a new chain.

- *Sieve 5 – Location match*

In this sieve, if two mentions are labeled with the "LOC" named-entity tag and one mention is a substring of another, they will be merged into a partial chain. For example, the two mentions "Tehran" and "Tehran city" are placed in the same chain using this sieve.

- *Sieve 6 – Title match*

In most Persian news texts, either the person's name comes first, then the title is mentioned, or the title comes first, and then the person's name comes. These two mentions (i.e., the person and its title) are merged in this sieve following simple rules. For example, the two mentions "President of France" and "Emmanuel Macron" are connected in this sieve.

- *Sieve 7 : demonstrative phrase match*

In this sieve, the reference of demonstrative expressions is determined. For example, if the text contains the mention "flower exhibition" and the mention "this exhibition" also appears, these two mentions will form a new chain through this sieve.

- *Sieve 8 : Pronoun resolution*

The sieves introduced up to this point form partial coreference chains. In this sieve, the pronoun mentions are connected to one of the previous partial chains based on the feature matching. Table 6 compares the proposed sieve-based system's features and the system [22].

Table 6. Features related to the pronoun sieve in the system [22] and the presented system

| Feature name | Proposed system | English sieve-based system |
|---|---|---|
| Number | 1- Static list for pronouns<br>2- NER labels<br>3- POS labels | 1- Static list for pronouns<br>2- NER labels<br>3- POS labels<br>4- Static dictionary from **[63]** |
| Gender | 1- Static lexicon<br>2- Static pronouns | 1- Static lexicon from **[63]** |
| Person | - | 1- Use for only pronouns |
| Animacy | 1- Static list for pronouns<br>2- NER labels<br>3- Static list for titles | 1- Static list for pronouns<br>2- NER labels<br>3- Dictionary extracted from WEB **[64]** |

| | 4- Static list for Persian names<br>5- Entity label from Persian corpus [۶۵] | |
|---|---|---|
| Distance | Less than 4 sentence | Less than 4 sentence |

## 4-2 Persian mention-pair baseline model

The second baseline is a mention-pair model based on hand-crafted features. In this model, pronoun, antecedent, and relational features have been developed according to the Persian language. A hyper-parameter called distance window is also defined. This hyper-parameter determines the maximum distance between the pronoun and its antecedent candidate.

In training, a positive example will be made between each pronoun and its closest real antecedent. In order to prevent an unbalanced dataset, the negative examples will include all mentions between the pronoun and its real antecedent [32]. After constructing the positive and negative samples, the feature vector is constructed.

Table 7 presents the feature vector related to the mention-pair and the proposed hybrid systems. As shown in the table, some features are at the mention level (M), and some features are at the cluster level (E). The sign "+" in the table means that this feature only belongs to the hybrid model presented in this article. The proposed hybrid system includes all the features in the table below. The mention-pair system includes features that do not have a sign "+." In order to choose the appropriate classifier, the results of 5 repetitions of 10-fold cross-validation have been checked on the training data and on the classifiers of the decision tree, random tree, multi-layer perceptron, and support vector machine. The best model on the mention-pair system is related to the random forest, which is given in the results section.

Table 7. The features used in the proposed hybrid models and the mention-pair model

| | Pronoun features | |
|---|---|---|
| row | Description | M/E |
| 1 | Is it a personal pronoun? | M |
| 2 | Is it a demonstrative pronoun? | M |
| 3 | Is it a reflexive pronoun? | M |
| 4 | Is it a third-person pronoun? | M |
| 5 | Is it a speech pronoun? | M |
| 6 | Pos label of the first token on the left side of the pronoun | M |
| 7 | Pos label of the second token on the left side of the pronoun | M |
| 8 | Pos label of the third token on the left side of the pronoun | M |
| 9 | Pos label of the first token on the right side of the pronoun | M |
| 10 | Pos label of the second token on the right side of the pronoun | M |
| 11 | Pos label of the third token on the right side of the pronoun | M |
| 12 | Is pronoun subject? | M |
| 13 | Is pronoun object? | M |
| 14 | Pronoun number | M |
| 15 | Pronoun animacy | M + |
| 16 | Pronoun person | M + |
| | Antecedent features | |
| 17 | How many tokens does the antecedent have? | M |
| 18 | Is the antecedent a pronoun? | M |
| 19 | Is the antecedent a demonstrative phrase? | M |
| 20 | Pos label of the first token on the left side of the antecedent | M |
| 21 | Pos label of the second token on the left side of the antecedent | M |
| 22 | Pos label of the third token on the left side of the antecedent | M |
| 23 | Pos label of the first token on the right side of the antecedent | M |
| 24 | Pos label of the second token on the right side of the antecedent | M |
| 25 | Pos label of the third token on the right side of the antecedent | M |
| 26 | Antecedent number | M |
| 27 | Is Antecedent subject? | M |
| 28 | Is Antecedent object? | M |
| 29 | How many mentions are there in the antecedent chain? | E+ |
| 30 | Is the antecedent a reflexive pronoun? | M+ |
| 31 | Antecedent type (pronoun, proper noun, common noun) | M+ |
| 32 | The sentence Index of the first mention in the antecedent chain | E+ |
| 33 | Antecedent animacy | M+ |
| 34 | Antecedent person | M+ |
| 35 | Named-entity label | M+ |
| 36 | Antecedent chain number | E+ |
| 37 | Antecedent chain animacy | E+ |
| | **Relational features** | |
| 38 | Sentence distance | M |
| 39 | Token distance | M |
| 40 | Number agreement | M |
| 41 | Subject agreement | M |
| 42 | Object agreement | M |
| 43 | String match | M |
| 44 | Is the distance between the pronoun and antecedent candidate less than three? | M |
| 45 | Are the pronoun and the antecedent candidate in the same sentence? | M |
| 46 | What is the minimum sentence distance between the pronoun and the antecedent candidate chain? | E+ |
| 47 | Is the antecedent string token count more than the pronoun string? | M+ |
| 48 | Animacy agreement | M+ |
| 49 | Person agreement | M+ |
| 50 | If a pronoun is an object and the antecedent is a subject, both are in the same sentence. | M+ |
| 51 | If a pronoun is reflexive and the antecedent candidate is a subject, both are in the same sentence. | M+ |
| | **Word embedding features** | |
| 52 | Euclidean distance of two mention heads | M+ |
| 53 | Euclidean distance between the mean of two mentions | M+ |
| 54 | The Euclidean distance between the antecedent candidate and the sentence in which the pronoun is placed. | M+ |
| - | Glove word embedding for pronoun | M+ |
| - | Glove word embedding for antecedent | M+ |

## 4-3 The proposed hybrid model

The previous two sections describe baseline rule-based and machine-learning models. In this section, a hybrid model has been used using the strengths of these two models. The hybrid system has the same results as end-to-end systems. This proposed hybrid model is as efficient as end-to-end models. One of the reasons for using the random forest in this research is that it has less computational complexity than deep learning models. The preprocessing tool first processes the corpus documents to form structured data. Then it will be a binary problem for each pronoun. Since the algorithm's input is structured data instead of raw textual data, using random forest has good results.

The base of the hybrid model is the sieve-base model described in Sections 4-1. This sieve is specially trained to identify the antecedent of the pronoun. The architecture of this model is shown in Figure **2** 2.

The entity-centric model aims to collect helpful information incrementally and involve this information in future decisions. The hybrid system presented in this paper consists of eight sieves. Seven rule-based sieves which used in the baseline model, and a machine-learning sieve is used for pronoun resolution. In the classification step on Mehr and RCDAT corpora, the best results were related to using random forest. Another reason is that random-forest can naturally model a large number of conjoined features.

In the results section, it will be shown that using the random forest in two corpora has better results than other algorithms, including multilayer perceptron. In this article, a hybrid model is used, which, in addition to its simplicity, can handle the challenges like end-to-end models. This model can provide interpretable output. In the following, the stages of model training and testing are described.

- o **Feature vector**

The appropriate selection of high-accuracy features is one of the essential steps in constructing pronoun resolution systems. Many knowledge sources are used for information extraction and feature vector construction. Most machine learning systems use a combination of these knowledge sources. The number of features used in various systems is different. The feature vector in the hybrid system complements the mention-pair baseline feature vector. Table7 shows 54 features used in the hybrid system. Also, glove word embedding has been added to the feature vector for pronouns and their antecedent candidates.

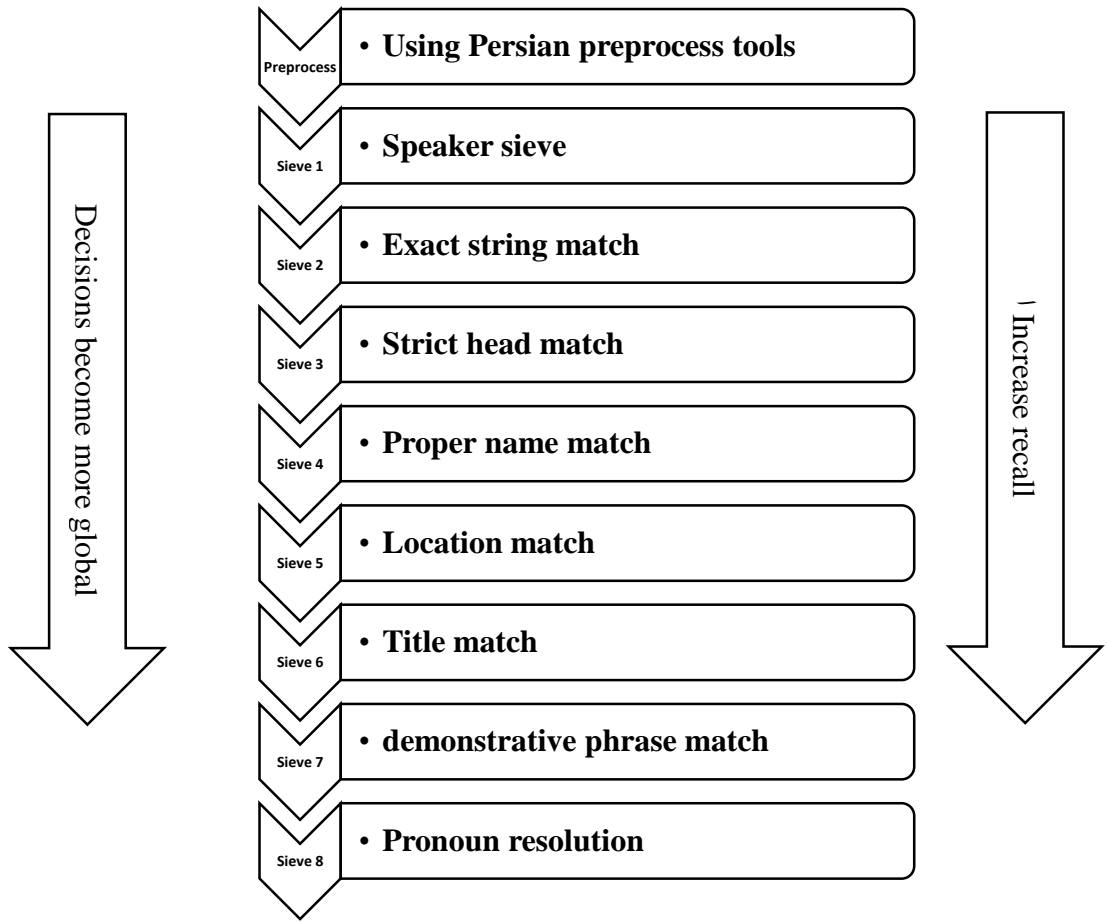

**Figure 2.** The introduced hybrid architecture.

- **Training**

In the entity-based model, the goal is to connect a noun phrase (in this research, a pronoun) to one of its previous chains. The pronouns machine learning sieve is trained on previously created chains. The method of training positive and negative examples is the same as the mention-pair model described in the previous section, with the difference that cluster-level features will be added to the model. Different classifiers are trained on two Persian corpora, the best results of which are given in the results section. For the classifier with the best results, the hyper-parameters are tuned in the development section, which is reported in the results section.

- **Test**

In this section, the Model finds an antecedent for each pronoun in the test data. First, partial entities are formed by applying rule-based sieves. Then, for each pronoun, the machine-learning sieve tries to connect this pronoun with all its previous partial entities. The antecedent candidate with the highest score from the random forest output will be selected as the antecedent. For each type of pronoun, a sentence window is considered. The random forest also uses a threshold limit estimated from the development section. If none of the partial clusters succeeds in passing this threshold, this pronoun will be non-anaphoric.

## 5. Experiments

To evaluate and compare the presented system with baseline systems and other systems presented in Persian, Mehr, and RCDAT corpora are used. The Mehr corpus trains on the Mehr training documents. Experiments on the Mehr corpus are performed on the Mehr test documents. The RCDAT corpus model trains on all training documents of this corpus. Because previous Persian systems that have implemented their model on the RCDAT corpus have used the Uppsala training documents [66] , in this research, the Uppsala corpus is used to test the model trained on the RCDAT corpus. The Uppsala corpus contains 415 sentences and 16274 tokens. This corpus consists of four separate parts that express four separate narratives.

This paper identifies the antecedent of detached pronouns in Persian. These pronouns include personal, demonstrative, and reflexive pronouns. Pronoun resolution is more challenging in Persian than in English. Here are some reasons for this difficulty. Persian is a null-subject language; nominal pronouns can be removed from the sentence. Also, in Persian, the third-person pronouns do not have a specific gender. That is, information about gender is ineffective in determining the antecedent of pronouns. Another case of complexity is that it is sometimes used to respect a plural pronoun for a singular noun group. Also, sometimes for respect, a plural pronoun is chosen as a noun phrase antecedent. Another reason is that sometimes a singular pronoun is used for a plural noun phrase.

### 5-1 Experiments on MEHR corpus

Table8 reports the results of Mehr corpus experiments and compares the proposed system with baseline and other Persian coreference systems. In this table, the results of the proposed system are reported for the best model parameters. Lines 1-4 report the results related to the baseline systems and the system presented in Persian with the highest efficiency. The authors of this article have compared their system with the most current English coreference systems, and their proposed system was more efficient in Persian.

Lines 5-9 are related to the results of the proposed hybrid system using random-forest with different settings. These settings include extraction type of cluster-level features (gold, automatic), using or not using glove word embedding for mentions, and using or not using rule-based sieves. Lines 1-4 report the results of baseline systems and the system presented in Persian with the highest efficiency. Lines 5-9 are related to the results of the proposed system using random forest. Line 5 shows the results of extracting gold cluster-level features. Lines 6 and 7 give different settings of the automatic proposed system. The gold features are extracted in lines 8-9, but the other two settings related to using rule-based sieves and glove word embedding have been changed. Lines 10-11 are the results of the proposed system using MLP with different settings. In line 12, the system results are reported by changing the position of the sieves. In line 13, the result of using linear regression instead of random forest in the system of line 7 is reported.

**Table 8**.The reported results of the presented system on Mehr corpus compared to previous works and various settings

|   | System | precision | recall | F1 |
|---|---|---|---|---|
| 1 | Rule-based baseline | 36.4 | 35.6 | 36 |

| | | | | | | |
|---|---|---|---|---|---|---|
| 2 | Mention-pair baseline (random forest) | | | 60 | 58.82 | 59.4 |
| 3 | The system proposed in [31] (random forest) | | | 55.08 | 54.54 | 55.08 |
| 4 | The system proposed in [31] (MLP) | | | 60.47 | 59.88 | 60.47 |
| | Proposed system with different settings (random forest) | | | | | |
| | **Cluster-based features** | **Using glove** | **Using rule-based sieves** | | | |
| 5 | gold | + | + | 66.56 | 66.27 | **65.90** |
| 6 | system | - | + | 64.2 | 62.94 | 63.56 |
| 7 | system | + | + | 65.4 | 64.11 | **64.75** |
| 8 | gold | - | - | 64.28 | 63.01 | 63.64 |
| 9 | gold | + | - | 66.28 | 64.98 | 65.62 |
| | Proposed system with different settings (MLP) | | | | | |
| | **Cluster-based features** | **Using glove** | **Using rule-based sieves** | | | |
| 10 | gold | + | + | 64.2 | 62.97 | 63.56 |
| 11 | system | + | + | 60.8 | 59.64 | 60.2 |
| 12 | Changing the sieves' order of system line 7 | | | 64.9 | 63.9 | 64.4 |
| 13 | Using logistic regression model instead of random forest in system line 7 | | | 60.42 | 58.44 | 59.41 |

### 5-1-1 Mehr corpus result analysis

Line 7 of table 8 shows the results related to the proposed system. This line shows that this system performs better than the previous best Persian system (an increase equal to 4.28 F1). This issue states that implementing a good hybrid architecture using cluster-level features makes it possible to provide good performance using the simplicity of the sieve-based model. The analysis of the table8 result is given below.

- o The rule-based baseline low accuracy indicates that using only rule-based sieves cannot be helpful in Persian pronoun resolution. The mention-pair machine-learning model shows a difference of 23 F1 scores compared to the rule-based model. The reason is that the random forest can correctly model the un-conjoined features in the pronoun resolution domain.
- o Line 7 reports the highest automatic hybrid model. This system uses rule-based sieves before applying the machine learning model. Rule-based sieves automatically form partial chains that, through them, the system can extract cluster-level features. In the feature set of this model, glove word embedding is used for pronouns and their antecedent mentions, which significantly impacts the model's efficiency.
- o The highest result of the report corresponds to line 5. Compared to the automatic system reported in line 7, gold cluster-level features are used in this system. The cluster-level features are obtained directly from the corpus in this setting. The difference of about 1 F1 score with the automatic system shows the importance of the accuracy of Persian preprocessing tools in rule-based sieves. This decreased accuracy indicates that the rule-based sieves do not form partial clusters correctly and have some errors.

- Considering the line 5 gold system and comparing it with the line 9 gold system in which the rule-based sieve is removed, it can be seen that the system's efficiency has decreased. This issue shows that some pronouns find their reference using rule-based sieves. So, removing these rule-based sieves reduces the system's efficiency. This issue shows the efficiency of using a hybrid architecture.
- By changing the sieves' order of the automatic system (result in line 12), only a 0.35 F1 score is deducted from the system's efficiency, showing that the order of sieves in the hybrid system is unimportant.
- Comparing the results of the proposed system with the previous Persian system of line 3 shows that adding cluster-level features and using hybrid architecture dramatically impacts the efficiency of the pronoun resolution system. The hybrid architecture provides cluster-level information using rule-based sieves. This information has a significant impact on the efficiency of the system.
- The linear logistic regression model is used in line 13 while maintaining the system architecture of line 7. The difference in the accuracy of these two systems is reported as about a 5 F1 score. This test shows the efficiency of random forest compared to a linear model. The reason for this is the ability of the random forest model to generate a non-linear model for un-conjoined features.
- The efficiency of using glove word embedding for pronoun and its antecedent candidate is another item deduced from the table. This word embedding has dramatically impacted the performance of both automatic and gold models.

*5-1-2 Hyper-parameter settings for implemented systems on Mehr corpus*

This section gives the hyper-parameters report obtained for different models on the Mehr corpus. The best sentence window for selecting the candidate antecedent in all the experiments is three. The 0.5 merging threshold to merge two mentions is used in all models. This threshold is tuned on the development set.

- **Mention-pair baseline**

A grid search shown in table 9 has been used to find the optimal hyper-parameters of the random forest.

Table 9. Random forest grid-search hyper-parameter values

| Hyper-parameter | Values |
|---|---|
| Max depth | None,10,20,30,40,50,60,70,80,90,100 |
| Number of estimators | 100,200,500,1000 |
| Criterion | Gini, Entropy |

The best-obtained hyper-parameters are 20, 1000, and Gini, respectively. The results of 10-fold cross-validation on the training data (without creating chains) using the best hyper-parameters are given in table 10.

Table10. The results of 10-fold cross-validation for the mention-pair model using the best hyper-parameters

| precision | recall | F1 |
|---|---|---|
| 85.65 | 64.46 | 73.52 |

- **Hybrid model**

To find the optimal parameters of the hybrid random forest model, a grid search with the values of table 9 has been used. The best hyper-parameters for the hybrid model were 30, 500, and Entropy, respectively. Table 11 shows the results of 10-fold cross-validation on the training data using the mentioned hyper-parameters.

**Table11.** 10-fold cross-validation results for the hybrid model using the best hyper-parameters

| | precision | recall | F1 |
|---|---|---|---|
| **Feature vector results without considering glove word embedding** | 88.71 | 81.33 | 84.85 |
| **Feature vector results with considering glove word embedding** | 91.17 | 79.63 | 84.96 |

A model with two hidden layers is used for the hybrid system using MLP instead of random forest. In the first layer, 1500 nodes are used with the relu activation function, and in the second layer, 300 nodes are used. Table 12 reports the best-adjusted hyper-parameters for this model.

**Table 12.** Optimal hyper-parameters of MLP hybrid model

| First hidden layer | |
|---|---|
| **Kernel initializer** | identity |
| **Regularization value** | 0.001 |
| **dropout** | 0.6 |
| **Second hidden layer** | |
| **Kernel initializer** | identity |
| **Regularization value** | 0.001 |
| **dropout** | 0.7 |
| **Other model settings** | |
| **Learning rates** | 0.0001 |
| **Optimization Alghorithm** | Adam |

The training results for this setting are reported in table 13.

**Table 13.** 10-fold cross-validation results for the hybrid MLP model using the best parameters

| precision | recall | F1 |
|---|---|---|
| 80.22 | 86.89 | 81.89 |

## 5-2 Experiments on RCDAT corpus

This section reports the results of the RCDAT corpus. Like Mehr corpus, the hybrid proposed system results are given on random forest and MLP classifiers. Different configurations of models and algorithms are reported in table14. In section 5-2-1, the analysis of the results of the experiments on the RCDAT is given. Then, in section 5-2-1, the hyper-parameters tuning is reported. The critical point in the results of the RCDAT corpus is that because, in this corpus, non-referential pronouns and pronouns at the beginning of a cluster are not labeled, in the results report, only the pronouns that are part of a chain are included in the evaluation process. The results report did not consider Singleton pronouns; only pronouns with a real antecedent participated in the final evaluation.

**Table 14.** The reported results of the presented system on RCDAT corpus compared to previous works and various settings

|   | System | | | precision | recall | F1 |
|---|---|---|---|---|---|---|
| 1 | Rule-based baseline | | | 42.66 | 41.15 | 41.89 |
| 2 | Mention-pair baseline (random forest) | | | 50.91 | 48.67 | 49.5 |
| 3 | The system proposed in [31] (random forest) | | | 43.66 | 42.2 | 42.88 |
| 4 | The system proposed in [31] (MLP) | | | 49.75 | 46.63 | **48.14** |
|   | Proposed system with different settings (random forest) | | | | | |
|   | **Cluster-based features** | **Using glove** | **Using rule-based sieves** | | | |
| 5 | system | - | - | 41.83 | 40.35 | 41.08 |
| 6 | system | + | - | 49.8 | 48.2 | 49 |
| 7 | system | + | + | 51.37 | 49.55 | **50.45** |
| 8 | gold | - | - | 53.76 | 51.85 | 52.79 |
| 9 | gold | + | - | 55.22 | 53.27 | 54.23 |
| 10 | gold | + | + | 56.05 | 54.07 | **55.04** |
|   | Proposed system with different settings (MLP) | | | | | |
|   | **Cluster-based features** | **Using glove** | **Using rule-based sieves** | | | |
| 11 | gold | + | + | 49.34 | 48.5 | 48.92 |
| 12 | system | + | + | 46.78 | 45.13 | 45.94 |

*5-2-1 RCDAT corpus result analysis*

According to line 7 of table14 , the presented system has increased the accuracy by about 2 F1 scores compared to the system [31] . According to row 9 of table14, if the cluster-level features are directly extracted from the corpus, this efficiency increase equals 7 F1. The analysis of the table 14 result is given below.

- The results of the rule-based system on the RCDAT corpus are disappointing as the results of the rule-based system on the Mehr corpus. A rule-based system alone cannot model un-conjoined features. Compared to the rule-based model, the mention-pair model results improve the model's efficiency.
- Line 7 of the table shows the results of the proposed system. This hybrid system first applies several rule-based sieves to build partial chains. Also, the cluster-level features are extracted from these partial chains. This model has an increase of 2 F1 scores compared to the most efficient model in the Persian language [31].
- The highest efficiency corresponds to line 10 of the table. This setting extracts cluster-level features directly from the corpus chain (gold chains). The increase in efficiency compared to the system in line 7 is that the rule-based sieves could not form partial chains accurately. This setting increases the accuracy by 7 F1 compared to the system [31].
- The higher efficiency of the proposed system using random forest compared to MLP indicates the ability of this classifier to model un-conjoined features. In all settings, the random forest model performs better than other classifiers.
- Comparing the results of lines 6-7 and 9- 10 proves the efficiency of using the outputs produced by rule-based sieves. Some pronouns are attached to their antecedent in rule-based sieves. For example, the speaker sieve may attach first-person pronouns to their antecedent. So, using the output of these sieves has increased the model's efficiency by about 1 F1 score.

*5-2-2 Hyper-parameter settings for implemented systems on RCDAT corpus*

Choosing the hyper-parameters of the models on the RCDAT corpus is given in this section. The sentence window is set to 3 for selecting antecedent candidates in all experiments. The merging threshold is set to 0.2 in all models. This hyper-parameter is tuned in the development section.

- **Mention-pair model**

Like the Mehr corpus, a grid search, according to table 9, has been used to find the optimal hyper-parameters of the random forest.

The best-obtained hyper-parameters are 40, 1000, and Entropy, respectively. The results of 10-fold cross-validation on the training data (without creating chains) using the best hyper-parameters are given in table 15.

Table15. The results of 10-fold cross-validation for the mention-pair model using the best hyper-parameters

| precision | recall | F1 |
|---|---|---|
| 84.19 | 58.46 | 69 |

- **Hybrid model**

The best hyper-parameters for the hybrid model, according to grid-search in table 9, are 50, 1000, and Entropy, respectively. Tableable 16 shows the results of 10-fold cross-validation on the RCDAT training data using the mentioned hyper-parameters.

Table16. 10-fold cross-validation results for the hybrid model using the best hyper-parameters.

|  | precision | recall | F1 |
|---|---|---|---|
| **Feature vector results without considering glove word embedding** | 88.75 | 81.48 | 84.95 |
| **Feature vector results with considering glove word embedding** | 88.79 | 81.46 | 84.96 |

A 3-layer MLP model has been used to train the hybrid model. The nodes in the layers are 850, 680, and 440, respectively. The node activation function is relu. Table 17 shows the best-adjusted hyper-parameters of the model.

Table 17. Optimal hyper-parameters of hybrid MLP model.

| First hidden layer | |
|---|---|
| **Kernel initializer** | glorot_normal |
| **Regularization value** | 0.001 |
| **dropout** | 0.5 |
| Second hidden layer | |
| **Kernel initializer** | identity |
| **Regularization value** | 0.01 |
| dropout | 0.1 |
| **Third hidden layer** | |
| **Kernel initializer** | identity |
| **Regularization value** | 0.001 |
| **dropout** | 0.1 |
| **Other model settings** | |
| **Learning rates** | 0.0001 |
| **Optimization Alghorithm** | Adam |

## 6. Conclusion

This article presents a hybrid model, combining the rule-based and machine-learning models for Persian pronoun resolution. In the presented method, a machine-learning component is provided to connect pronouns to their partial chains, which has increased the system's efficiency. In the rule-based component, seven sieves have been used in Persian, arranged in order of decreasing precision. A machine learning module is developed using random forest and cluster-level features to identify the antecedent of pronouns. One of the reasons for the better results of the random forest than other models is the ability of the random forest to model un-conjoined features.

One of the article's contributions is the development of a Persian coreference resolution corpus. This corpus contains 400 political, sports, cultural, and other news documents. In this corpus, non-referential pronouns and named entities within a noun phrase (nested noun phrase) are also labeled, unlike the previous corpora in Persian. The results have been published on two corpora, Mehr and RCDAT. In this paper, all the challenges in end-to-end models have been solved using a hybrid model, processing pipeline, and correctly using cluster-level features. The introduced model is simple, modular, interpretable, and based on the entity-centric model.

In the results, this system using the random forest model has had better results than the previous systems in Persian on both Mehr and RCDAT corpora. In the results report, all kinds of baseline models and different settings of the model and sieves are reported. Also, hyper-parameters tuning of different models is given in a separate section. The best Persian coreference resolution system reported its model's effectiveness by implementing and comparing the latest English systems on the Persian RCDAT corpus. This system has an increase of 4.28 F1 score on Mehr corpus and an increase of about 2 F1 scores on RCDAT compared to the best Persian system. Developing an end-to-end model using transfer learning can be a good choice for future work.